\pdfoutput=1
\documentclass[11pt]{article}

\usepackage[final]{acl}
\usepackage[]{acl}

\usepackage{times}
\usepackage{latexsym}

\usepackage[T1]{fontenc}
\usepackage[utf8]{inputenc}

\usepackage{microtype}

\usepackage{hyperref}
\usepackage{amsmath} % 公式包
\usepackage{amssymb} % 公式包
\usepackage{graphicx} % 加载图像
 \usepackage{xcolor}
\usepackage{threeparttable}
\usepackage{pifont}
\usepackage{multirow}
\usepackage{subfigure}
\usepackage{graphicx}
\usepackage{booktabs}
\usepackage{enumitem}
\usepackage{algorithm}
\usepackage{algpseudocode}
\usepackage{amsmath}
\usepackage{graphicx} % 引入graphicx包来使用\includegraphics命令
\usepackage{mdframed}
\usepackage{relsize}
\usepackage{ulem}
\usepackage{booktabs}
\usepackage{graphicx}
\usepackage{array}
\usepackage{colortbl} % Include the colortbl package for row coloring
\usepackage{array}    % Allows column spacing control

 \usepackage{amsmath}

\usepackage{csquotes}
\usepackage{epigraph}
\usepackage{multirow} 
\usepackage{geometry}
\usepackage{array}
\usepackage{longtable}
\usepackage{float}
\usepackage{mdframed}
\usepackage{xcolor}
\definecolor{deepyellow}{rgb}{0.8, 0.7, 0.0} 
\definecolor{deepgreen}{rgb}{0.0, 0.7, 0.0} 

\definecolor{chatgpt_c}{RGB}{121,147,210}
\definecolor{src_c}{RGB}{238,154,189}
\definecolor{tgt_c}{RGB}{139,212,209}
\definecolor{src_tgt_c}{RGB}{149,149,149}
\definecolor{ibut_c}{RGB}{247,182,95}

\definecolor{chinese_red}{RGB}{230,239,255}
\definecolor{chinese_red_small}{RGB}{252,255,230}
\definecolor{chinese_brown}{RGB}{246,230,255}
 \usepackage{booktabs}

\definecolor{win}{RGB}{165,127,183} % 定义颜色
\definecolor{tie}{RGB}{204,161,189}
\definecolor{loss}{RGB}{229,207,221}

\usepackage{CJK}

\title{LLM-based Translation Inference with Iterative Bilingual Understanding}

\author{
Andong Chen\thanks{Work was done when Andong Chen was at Pengcheng Laboratory.}$^{1}$\hspace{0.5mm}, 
 Lianzhang Lou$^{2}$\hspace{0.5mm}, 
\textbf{Kehai Chen}\thanks{~~Corresponding author.}$^{1,2}$\hspace{0.5mm}, 
 \textbf{Xuefeng Bai}$^{1}$\hspace{0.5mm}, 
 \textbf{Yang Xiang}$^{2}$\hspace{0.5mm}, \\
 \textbf{Muyun Yang}$^{1}$\hspace{0.5mm},
 \textbf{Yang Feng}$^{3}$\hspace{0.5mm}
 \textbf{Tiejun Zhao}$^{1}$\hspace{0.5mm}, 
 \textbf{Min Zhang}$^{1}$\hspace{0.2mm}\hspace{1.5mm} \\
$^1$ School of Computer Science and Technology, Harbin Institute of Technology, China\\
$^2$ Pengcheng Laboratory, Shenzhen, China \\
$^3$ Key Laboratory of Intelligent Information Processing, \\
Institute of Computing Technology, Chinese Academy of Sciences (ICT/CAS) \\
  ands691119@gmail.com,  \{loulzh, xiangy\}@pcl.ac.cn, fengyang@ict.ac.cn,  \\
  \{chenkehai, baixuefeng, yangmuyun, tjzhao, zhangmin2021\}@hit.edu.cn   
}

\begin{document}
\maketitle
\begin{abstract}
The remarkable understanding and generation capabilities of large language models (LLMs) have greatly improved the performance of machine translation.
However, a poor understanding often leads to the misleading of key information within one input sentence (e.g., concepts and terms), called \textbf{\textit{understanding distortion}}, thereby degrading the quality of target language translations generated by LLMs. To alleviate this issue, we propose a novel Iterative Bilingual Understanding Translation (IBUT) method to enhance the understanding of sentences. Particularly, IBUT explicitly generates the contextual understanding of source and target sentences explaining key concepts, terms, examples, etc. Thus, IBUT utilizes the dual characteristics of machine translation to generate effective cross-lingual feedback, and thereby iteratively refines contextual understanding to improve the translation performance of LLMs. Experimental results showed that the proposed IBUT significantly outperforms several strong comparison methods on the multiple domain benchmarks (e.g., news, commonsense, and cultural). Source codes will be released.

% The remarkable understanding and generation capabilities of large language models (LLMs) have greatly improved translation performance. However, incorrect understanding of the sentence to be translated can degrade translation quality.
% To address this issue, we propose a novel Iterative Bilingual Understanding Translation (IBUT) method based on the cross-lingual capabilities of LLMs and the dual characteristics of translation tasks.
% The cross-lingual capability of LLMs enables the generation of contextual understanding for both the source and target languages separately. Furthermore, the dual characteristics allow IBUT to generate effective cross-lingual feedback, iteratively refining contextual understanding, thereby reducing errors and improving translation performance.
% Experimental results showed that the proposed IBUT outperforms several strong comparison methods, especially being generalized to multiple domains (e.g., news, commonsense, and cultural translation benchmarks). Our dataset and code will be made available.

% Our code will be made available at \url{https://github.com/anonymous}.
%Further analysis shows that the noise in contextual understanding is correlated with translation performance, and the IBUT method can effectively reduce this noise, thereby improving translation quality.

\end{abstract}


\section{Introduction}

% In the field of machine translation (MT), translations based on LLMs (LLM-MT) have become a research focus\citep{tyen2023llms,liang2023encouraging,DBLP:journals/corr/abs-2303-16104,DBLP:journals/corr/abs-2308-14186}. Currently, the use of LLMs to generate source or target-side contextual understanding \cite{bubeck2023sparks,xu2023large,zhao2023survey,shinn2023reflexion} (Figure \ref{intro} (b)) has demonstrated remarkable improvements in translation quality \citep{ouyang2022training,moslem2023adaptive,chen2023iterative,He2023ExploringHT,liang2023encouraging,DBLP:journals/corr/abs-2302-07856,DBLP:journals/corr/abs-2305-06575}. 

Large language models (LLMs) have shown impressive performance across multilingual machine translation~\cite{tyen2023llms,liang2023encouraging,DBLP:journals/corr/abs-2303-16104,DBLP:journals/corr/abs-2308-14186,zhang-etal-2024-paying}. 
%In the field of machine translation (MT), translations based on LLMs (LLM-MT) have become a research focus
Particularly, the remarkable understanding and generation capabilities of LLMs have greatly improved the translation performance~\cite{hendy2023good,jiao2023chatgpt,le2023bloom,DBLP:conf/wmt/IyerCB23,DBLP:journals/corr/abs-2311-02851,zhao2024review}. 
\begin{figure}[!th]\centering
\centering
\includegraphics[scale=0.5]{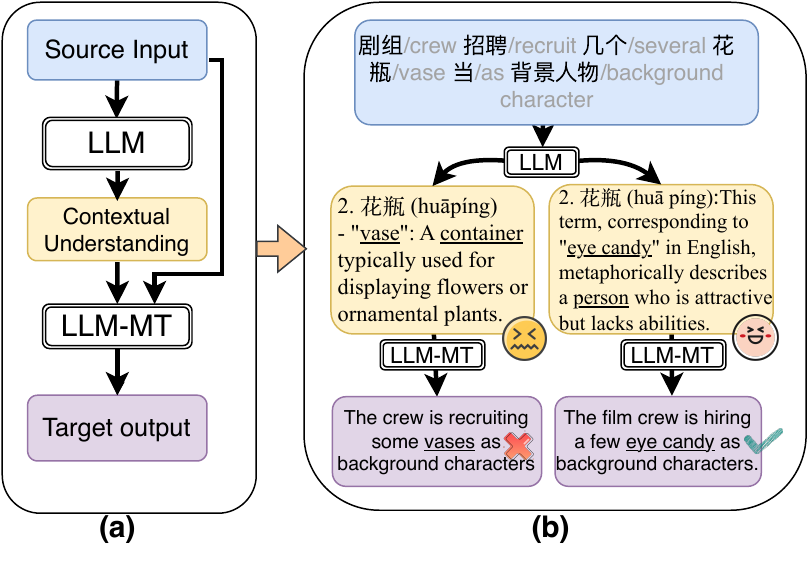} 
\caption{Illustration of the LLMs translation paradigm based on contextual understanding (Fig a). A commonsense domain example of LLM (gpt-3.5-turbo) translation from Chinese to English (Fig b).}
\label{intro}
\end{figure}
Typically, the LLM-based translation paradigm~\cite{he-etal-2024-exploring,chen-etal-2024-dual,liang2023encouraging,wu2024beyond,chen2024benchmarking} (as shown in Figure \ref{intro}(a)) first generates a contextual understanding of the sentence to be translated, for example the explanations of key concepts, terms or examples. 
Thus, this contextual understanding is used to help LLMs understand the key information of the input sentence, thereby enhancing the translation performance of LLMs.

However, when the LLM generates a poor contextual understanding of the input sentence, there is often a misleading of key information within one input sentence (e.g., concepts and terms), called \textbf{\textit{Understanding Distortion}} as shown in Figure 1 (b), thereby degrading the generation quality of the target language translation. 
% 这里结合Figure 1 (b)分析展示译文中的错误，尤其是对关键概念或者术语的错误译文，这样才会和下面的内容对应起来。
For example, in Figure 1(b), the LLM incorrectly understands "\begin{CJK*}{UTF8}{gbsn}花瓶/vase\end{CJK*}" as "a container for arranging flowers," resulting in a commonsense error in the translation output.
As a result, we think that Understanding Distortion may heavily hinder the translation advancement of LLMs.
%These generated understanding errors lead to the introduction of misleading information during the translation process, particularly when dealing with complex concepts such as commonsense and cultural domains. 
%We refer to this issue as \textbf{Understanding Distortion} of LLMs. This study manually evaluated the Chinese-English test set in the commonsense domain. It found that Understanding Distortion makes up \textbf{40\% }of translation errors, highlighting the importance of this issue (\textsection \ref{human_eval_error}).

To alleviate this issue, we propose a novel \textbf{I}terative \textbf{B}ilingual \textbf{U}nderstanding \textbf{T}ranslation (IBUT) approach to the contextual understanding of sentences in LLMs. 
To this end, IBUT consists of four parts: 1) Bilingual Understanding Generation leverages the cross-linguistic capabilities of LLMs to generate contextual understanding for both the source and target languages; 2) Alignment Judgment uses the generated bilingual contextual understanding and employs dual learning from the translation task \cite{he2016dual,qin2020dual,chen-etal-2024-dual} as supervisory signals to produce explicit verbal feedback;
% IBUT uses LLMs to generate bilingual contextual understanding. For the Alignment Judgment part, IBUT leverages the dual learning\cite{he2016dual,qin2020dual,chen-etal-2024-dual} of the translation task as supervisory signals to generate explicit verbal feedback. 
3) Iterative Refinement takes the verbal feedback as a $semantic$ gradient, providing LLM with a clear direction for refinement, thereby iteratively refining the bilingual contextual understanding; 4) Undersderstanding-Based Translation guides LLMs to generate the final translation depending on the input sentence and refined bilingual contextual understanding.

Experimental results showed that the proposed IBUT significantly outperforms several strong closed-source and open-source LLMs (including ChatGPT, GPT-4, Alpaca, and Vicuna), on the multiple domains (e.g., news, commonsense, and cultural) benchmarks. 
%The results show an average improvement of +1.3, +4.2, and +2.3 COMET scores compared to the baseline, confirming the effectiveness of the IBUT strategy. 
Additionally, quantitative and qualitative analyses show that IBUT helps LLMs learn a better contextual understanding, thereby improving translation performance. 

\section{Related Work}
\textbf{Machine Translation Based on Large Language Models (LLM-MT)}. Large language models, such as GPT-3 \cite{brown2020language}, have demonstrated their effectiveness in machine translation across various language pairs \cite{hendy2023good,jiao2023chatgpt,le2023bloom,DBLP:conf/wmt/IyerCB23,DBLP:journals/corr/abs-2311-02851,DBLP:conf/wmt/KarpinskaI23,DBLP:conf/wmt/MoslemRMKHW23,DBLP:conf/emnlp/WangLJZY0T23,DBLP:conf/wmt/IyerCB23,DBLP:conf/emnlp/FarinhasSM23}. Recent studies delve into the performance of LLM in machine translation, including control over formality in translation outputs \cite{garcia2022using}, in-context translation abilities during pre-training \cite{shin2022effect}, and the impact of LLM-based machine translation on culturally sensitive texts  \cite{DBLP:journals/corr/abs-2305-14328}.Additionally, a study has explored the bilingual capabilities of LLMs to enhance translation performance \cite{huang2024aligning}. For translation tasks requiring reasoning, multi-agent debates can effectively enhance the reasoning abilities of LLM-MT \cite{liang2023encouraging}. These investigations further validate the research value of LLM-MT, offering diverse research directions for scholars.

\textbf{Knowledge-based Machine Translation}. Extensive research indicates that incorporating knowledge enhances translation performance. This external knowledge includes bilingual dictionaries\cite{arthur2016incorporating}, probabilistic interpolation of dictionaries\cite{khandelwal2020nearest}, data augmentation through back-translation \cite{hu2019domain}, and entity-based denoising pre-training \cite{hu2021deep}. Additionally, researchers introduced domain \cite{gao2023design} and part-of-speech information during the inference phase and obtained multilingual translations of key terms through the NLLB translator \cite{lu2023chain}, thereby enhancing the translation quality for low-resource languages. LLMs improve MT by integrating internal knowledge like keywords, themes, and examples from source sentences \cite{He2023ExploringHT}. LLMs enhance MT performance by generating sentence-level understanding \cite{huang2024aligning,chen2024benchmarking}.

\section{Iterative Bilingual Understanding Translation}

\label{proposed_method}

\begin{figure*}[!th]\centering
\includegraphics[scale=0.190]{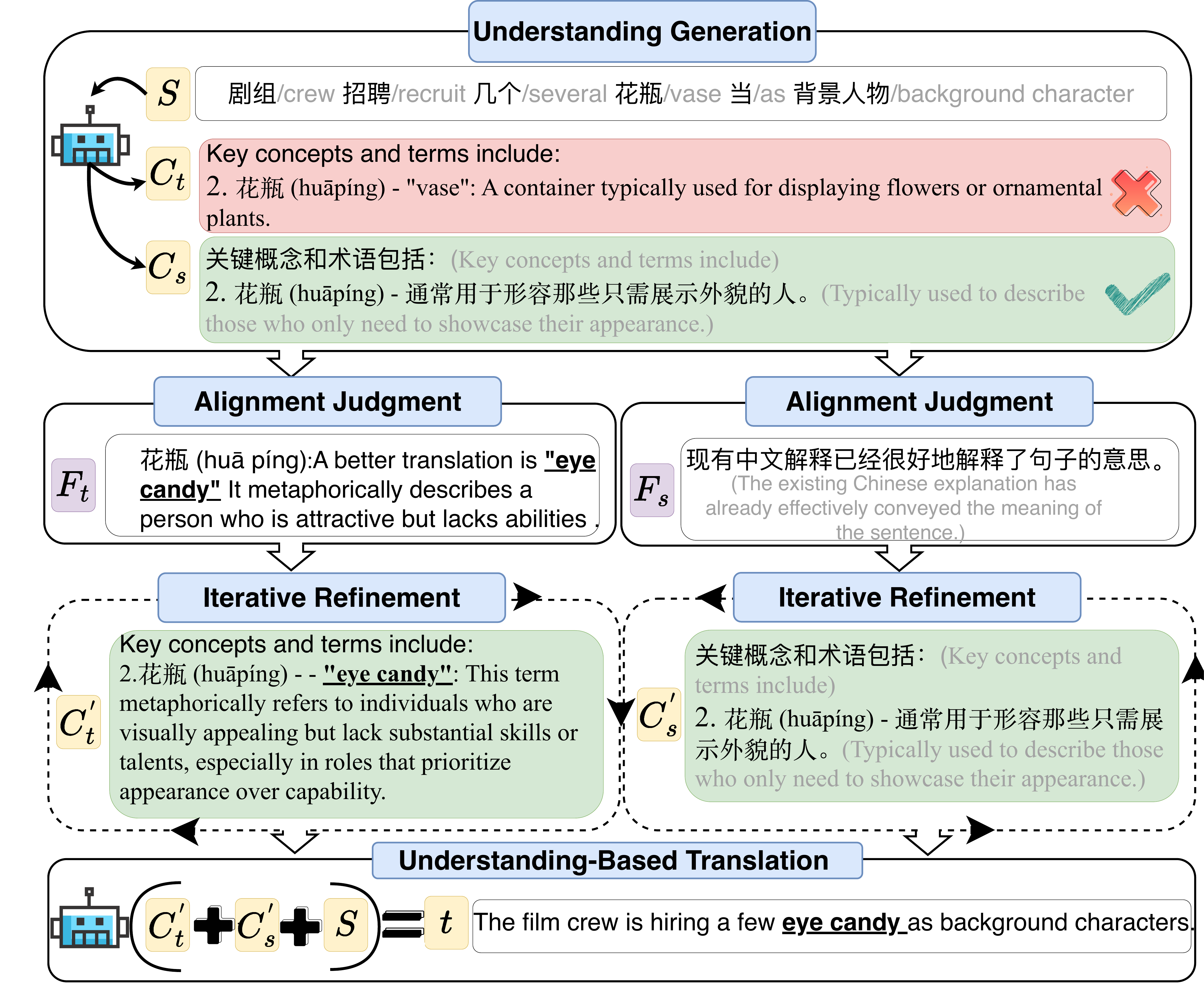} 
\caption{\textbf{IBUT translation framework. }The process involves first generating a bilingual understanding of the translation input sentence using an LLM. Next, verbal feedback is obtained via LLM, informed by the translation input and the bilingual understanding. This feedback is then used to further refine the bilingual understanding. The final step involves using LLM to perform the translation, leveraging both the bilingual understanding and the original input sentence. \textcolor{gray}{Gray text} indicates English annotations for the Chinese.}
\label{method}
\end{figure*}

The poor understanding of translated sentences generated by LLMs leads to a decline in translation quality. To address this issue, we propose a new method called \textbf{I}terative \textbf{B}ilingual \textbf{U}nderstanding \textbf{T}ranslation (IBUT). IBUT utilizes LLMs to generate bilingual contextual understanding of the source input and utilizes the dual learning of translation tasks to establish verbal feedback for iteratively refining this understanding. Finally, the iterative refinement reduces errors in bilingual contextual understanding, thereby enhancing translation performance. The IBUT consists of four parts: 1) Understanding Generation; 2) Alignment Judgment; 3) Iterative Refinement; 4) Understanding-Based Translation. We use \( MT \) to denote a translation model based on LLM, and lowercase letters \( s \) and \( t \) to represent sentences in the source language (\(L^s\)) and target language (\(L^t\)), respectively. That is, \( s = (s[1], \cdots, s[n]) \) and \( t = (t[1], \cdots, t[m]) \), where each \( s[i] \) and \( t[i] \) is a token.

\textbf{Understanding Generation.} For the first part of the IBUT method, as shown in Figure \ref{method}, LLMs generate contextual understanding in both the source and target languages from the source sentence, represented as \(C_s\) and \(C_t\) respectively. This understanding includes key concepts, terms, term explanations, and examples. Detailed prompts are provided in Appendix \ref{detail stage1}.

% 这里是左侧文本
\textbf{Alignment Judgment.} The second part of IBUT introduces an LLM-based agent, denoted as \(JA\), which evaluates the consistency of bilingual contextual understanding and supervises the entire translation process. Based on the dual learning \cite{he2016dual,qin2020dual,chen-etal-2024-dual}, bilingual contextual understanding is generated from the same source sentence, and both should be consistent in form and semantics. Based on this assumption, \(JA\) initially identifies whether there are differences in the bilingual contextual understanding (\(C_s\) and \(C_t\)) generated based on the source sentence \(s\). If \(JA(C_s, C_t, s) = \text{True}\), as shown in Figure \ref{method}, \(JA\) generates explicit verbal feedback in both the source and target languages (\(F_s, F_t \leftarrow \text{JA}(C_s, C_t, s)\)). The verbal feedback specifies the content of the differences between \(C_s\) and \(C_t\) and provides suggestions for refinement. If \(JA(C_s, C_t, s) = \text{False}\),  the process moves to Understanding-Based Translation (Appendix \ref{detail stage2} for prompts).

\textbf{Iterative Refinement.} In the third part of IBUT, the max number of iterations (\(max\_iter\)) is initially defined. As shown in Figure \ref{method}, the previously generated bilingual contextual understanding is refined based on the verbal feedback signals \(F_s\) and \(F_t\) (\(C^{'}_s \leftarrow M(s, C_s, F_s)\) and \(C^{'}_t \leftarrow M(s, C_t, F_t)\), where \(M\) is an LLM). If the number of iterations exceeds \(max\_iter\), the process will directly enter the Understanding-Based translation part. If the number of iterations does not exceed \(max\_iter\), the process will continue into the Alignment Judgment part to iteratively refine the bilingual contextual understanding.  Specific prompts are displayed in Appendix \ref{detail stage3}.

\textbf{Understanding-Based Translation.} In the final part of IBUT, the refined bilingual contextual understanding ($C^{'}_s$ and $C^{'}_t$) and the sentence to be translated are taken as inputs, and the translation is directly carried out through LLM-MT (\( t = MT(s, C^{'}_s, C^{'}_t \)). See Appendix \ref{detail stage4} for prompts.

\begin{table*}[!ht] \centering 
\scalebox{0.85}{ 
% \footnotesize % 使用较小的字体大小
\setlength{\tabcolsep}{3.3pt} % 减少列之间的距离
\begin{tabular}{lclclcclccclcc}
\hline
\multicolumn{1}{c}{\textbf{WMT22}}              & \multicolumn{2}{c}{\textbf{En$\rightarrow$De}} & \multicolumn{2}{c}{\textbf{En$\rightarrow$Ja}} & \textbf{Cs$\rightarrow$Uk} & \multicolumn{2}{c}{\textbf{Uk$\rightarrow$Cs}} & \textbf{En$\rightarrow$Hr} & \multicolumn{1}{l}{\textbf{Sah$\rightarrow$Ru}} & \multicolumn{2}{c}{\textbf{Ru$\rightarrow$Sah}} & \textbf{En$\rightarrow$Liv} & \multicolumn{1}{l}{\textbf{Liv$\rightarrow$En}} \\ \hline

\rowcolor{gray!30} 
\multicolumn{1}{c}{}                     & \multicolumn{13}{c}{COMET $\uparrow$ / BLEURT $\uparrow$}                                                                                                                                                                                                                                                                                                                                                         \\ \hline
ChatGPT                                         & \multicolumn{2}{c}{85.8/75.6}             & \multicolumn{2}{c}{88.3/66.3}             & 89.7/79.0            & \multicolumn{2}{c}{88.7/79.0}             & 86.6/76.8             & 57.5/36.0                                   & \multicolumn{2}{c}{52.8/73.2}               & 52.7/41.8            & 40.6/39.0                                \\
\quad+5shot                      & \multicolumn{2}{c}{86.5/76.3 }             & \multicolumn{2}{c}{88.2/67.1 }             & 88.3/75.6              & \multicolumn{2}{c}{89.6/79.1 }             & 86.4/75.6             & 58.3/36.0                                    & \multicolumn{2}{c}{53.1/75.4 }               & 55.3/42.1                & 42.7/40.9                                   \\
\quad+Rerank & \multicolumn{2}{c}{86.2/75.3 }             & \multicolumn{2}{c}{88.0/66.6 }             & 88.3/75.3              & \multicolumn{2}{c}{89.7/79.5 }             & 86.3/75.4              & 58.6/36.3                                    & \multicolumn{2}{c}{53.8/75.9 }               & 55.5/42.7                & 42.9/41.0                                   \\
\quad+MAD                        & \multicolumn{2}{c}{86.5/76.4 }             & \multicolumn{2}{c}{88.4/67.9 }             & 90.2/79.3              & \multicolumn{2}{c}{89.6/79.3 }             & 87.0/76.9              & 58.1/37.1                                    & \multicolumn{2}{c}{53.5/76.4 }               & 55.5/42.5                & 43.2/\textbf{41.3}                                   \\
\quad+MAPS                       & \multicolumn{2}{c}{86.4/76.3 }             & \multicolumn{2}{c}{88.5/67.4 }             & 88.8/76.1              & \multicolumn{2}{c}{89.8/79.6 }             & 86.5/76.0             & 58.7/37.3                                    & \multicolumn{2}{c}{53.3/76.1 }               & 54.1/42.0                & 43.6/39.7                                   \\
\quad+Refine                     & \multicolumn{2}{c}{86.0/75.9 }             & \multicolumn{2}{c}{88.6/67.9 }             & 89.8/79.0              & \multicolumn{2}{c}{89.3/79.8 }             & 87.0/76.9              & 58.3/37.4                                    & \multicolumn{2}{c}{53.8/76.5}                & 55.5/42.7                & 43.9/40.1                                   \\
\quad+TEaR                     & \multicolumn{2}{c}{86.2/76.2}             & \multicolumn{2}{c}{88.0/67.3}             & 88.7/77.3             & \multicolumn{2}{c}{89.3/79.2}             & 87.2 /76.2            & 58.3/37.2                                   & \multicolumn{2}{c}{53.4/75.3}               & 54.7/42.9              & 43.5/  39.8                                 \\
\quad+Dual-Reflect                    & \multicolumn{2}{c}{85.8/75.1}             & \multicolumn{2}{c}{88.3/67.2}             & 88.9/76.3             & \multicolumn{2}{c}{87.1/79.0}             & 58.2/76.9             & 58.0/37.1                                   & \multicolumn{2}{c}{58.2/74.2}               & 53.7/43.0             & 43.1/38.1                                  \\
\quad+IBUT                       & \multicolumn{2}{c}{\textbf{87.0/77.0 }}    & \multicolumn{2}{c}{\textbf{89.5/69.9 }}    & \textbf{91.2/80.1 }    & \multicolumn{2}{c}{\textbf{90.0/80.1 }}    & \textbf{87.8/77.1 }    & \textbf{59.5/37.9 }                          & \multicolumn{2}{c}{\textbf{54.5/76.9 }}      & \textbf{56.7/44.2 }      & \textbf{47.1}/40.5                         \\ \hline

\rowcolor{gray!30} 
\multicolumn{1}{c}{}                     & \multicolumn{13}{c}{BLEU $\uparrow$}  \\ \hline
ChatGPT                                         & \multicolumn{2}{c}{32.3}             & \multicolumn{2}{c}{17.3}             & 29.9             & \multicolumn{2}{c}{30.6}             & 26.9             & 5.9                                   & \multicolumn{2}{c}{1.9}               & 2.4               & 8.5                                   \\
\quad+5shot                      & \multicolumn{2}{c}{32.9}             & \multicolumn{2}{c}{17.9}             & 29.3             & \multicolumn{2}{c}{31.2}             & 25.8             & 6.4                                   & \multicolumn{2}{c}{2.3}               & 2.7               & 8.8                                   \\
\quad+Rerank & \multicolumn{2}{c}{33.6}             & \multicolumn{2}{c}{21.2}             & 29.5             & \multicolumn{2}{c}{31.9}             & 26.9             & 6.5                                   & \multicolumn{2}{c}{2.6}               & 2.9               & 8.9                                   \\
\quad+MAD                        & \multicolumn{2}{c}{32.9}             & \multicolumn{2}{c}{19.7}             & 31.6             & \multicolumn{2}{c}{31.6}             & 26.5             & 6.7                                   & \multicolumn{2}{c}{2.6}               & 3.1               & 9.7                                   \\
\quad+MAPS                       & \multicolumn{2}{c}{33.1}             & \multicolumn{2}{c}{21.2}             & 29.5             & \multicolumn{2}{c}{31.4}             & 27.0             & 6.7                                   & \multicolumn{2}{c}{2.2}               & 2.9               & 9.7                                   \\
\quad+Refine                     & \multicolumn{2}{c}{33.8}             & \multicolumn{2}{c}{23.4}             & 30.3             & \multicolumn{2}{c}{32.8}             & 27.5             & 6.7                                   & \multicolumn{2}{c}{2.5}               & 3.3               & 9.5                                   \\
\quad+TEaR                     & \multicolumn{2}{c}{33.8}             & \multicolumn{2}{c}{23.4}             & 30.3             & \multicolumn{2}{c}{32.8}             & 27.5             & 6.7                                   & \multicolumn{2}{c}{2.5}               & 3.3               & 9.5                                   \\
\quad+Dual-Reflect                     & \multicolumn{2}{c}{32.4}             & \multicolumn{2}{c}{20.2}             & 29.4             & \multicolumn{2}{c}{31.9}             & 26.4             & 6.5                                   & \multicolumn{2}{c}{2.6}               & 3.2              & 9.4                                   \\
\quad+IBUT                       & \multicolumn{2}{c}{\textbf{34.5}}    & \multicolumn{2}{c}{\textbf{24.3}}    & \textbf{31.9}    & \multicolumn{2}{c}{\textbf{34.3}}    & \textbf{28.5}    & \textbf{6.9}                          & \multicolumn{2}{c}{\textbf{4.9}}      & \textbf{4.7}      & \textbf{10.1}  \\ \hline
\end{tabular}}
\caption{The main results from the WMT22 news benchmark are presented. ChatGPT mean to perform translation directly through Zero-Shot. The bold indicates the highest scores that are statistically significant, with p-values less than 0.01 in the paired t-test against all compared methods.}
\label{wmt22test}
\end{table*}

\section{Experimental Setup}

\label{setup}
\textbf{Dataset}: We conduct experiments on four MT benchmarks: WMT22, WMT23 (general news MT benchmarks), commonsense MT, and cultural MT. Dataset details are in Appendix \ref{data_appendix}.

\begin{table*}[!ht]
\centering
\scalebox{0.77}{ 
\setlength{\tabcolsep}{3.3pt} % 减少列之间的距离
\begin{tabular}{lcllcllcllcccll}
\hline
\textbf{Culture}                     & \multicolumn{3}{c}{\textbf{En$\rightarrow$Es}}                                                  & \multicolumn{3}{c}{\textbf{En$\rightarrow$Fr}}                                                  & \multicolumn{3}{c}{\textbf{En$\rightarrow$Hi}}       & \textbf{En$\rightarrow$Ta}      & \multicolumn{1}{c}{\textbf{}\textbf{En$\rightarrow$Te}} & \multicolumn{3}{c}{\textbf{En$\rightarrow$Zh}}       \\ \hline
\rowcolor{gray!30} 
                     & \multicolumn{14}{c}{COMET $\uparrow$ /BLEURT $\uparrow$ /BLEU $\uparrow$}                                                                                                                                                                                                                                                                                                       \\ \hline
ChatGPT                     & \multicolumn{3}{c}{83.0 / 69.3 / 35.7}                                                     & \multicolumn{3}{c}{77.9 / 58.3 / 31.1}                                                     & \multicolumn{3}{c}{73.6 / 61.8 / 18.8}          & 67.9 / 57.4 / 11.3          & 69.9 / 52.0 / 13.2                         & \multicolumn{3}{c}{83.3 / 64.5 / 35.0}          \\
\quad+5-shot & \multicolumn{3}{c}{83.2 / 70.3 / 44.0}                                                     & \multicolumn{3}{c}{78.0 / 58.5 / 24.0}                                                     & \multicolumn{3}{c}{74.3 / 63.7 / 18.7}          & 71.8 / 60.2 / 11.2          & 70.6 / 53.6 / 13.3                         & \multicolumn{3}{c}{83.2 / 64.9 / 35.3}          \\
\quad+Rerank & \multicolumn{3}{c}{82.7 / 70.5 / 43.9}                                                     & \multicolumn{3}{c}{78.1 / 58.2 / 24.5}                                                     & \multicolumn{3}{c}{73.9 / 62.4 / 18.8}          & 70.5 / 59.4 / 11.2          & 70.4 / 52.7 / 13.0                         & \multicolumn{3}{c}{83.0 / 64.6 / 34.4}           \\
\quad+MAD    & \multicolumn{3}{c}{83.4 / \textbf{70.8} / 43.8}                           & \multicolumn{3}{c}{78.5 / \textbf{59.0} / 31.0}                           & \multicolumn{3}{c}{71.6 / 60.5 / 18.1}          & 71.1 / 60.3 / 11.5          & 71.0 / 53.6 / 13.3                         & \multicolumn{3}{c}{83.6 / 64.7 / 34.5}          \\
\quad+MAPS   & \multicolumn{3}{c}{82.9 / 70.0 / 42.1}                                                     & \multicolumn{3}{c}{78.2 / 58.7 / 30.6}                                                     & \multicolumn{3}{c}{71.8 / 60.4 / 11.9}          & 72.1 / 60.7 / 11.2          & 72.0 / 54.8 / 13.6                         & \multicolumn{3}{c}{83.5 / 64.1 / 34.6}          \\
\quad+Refine & \multicolumn{3}{c}{83.0 / 70.1 / 42.1}                                                     & \multicolumn{3}{c}{78.0 / 58.3 / 30.4}                                                     & \multicolumn{3}{c}{74.3 / 63.2 / 18.8}          & 71.8 / 60.9 / 11.7          & 71.7 / 54.6 / 13.7                         & \multicolumn{3}{c}{83.0 / 65.1 / 34.7}          \\
\quad+TEaR   & \multicolumn{3}{c}{82.6 / 70.3 /  43.3}                                                     & \multicolumn{3}{c}{77.1 / 58.7 / 30.2}                                                     & \multicolumn{3}{c}{71.4 / 61.2 / 15.3}          & 71.9 / 59.3 / 10.5          & 71.7 / 53.4 / 12.8                         & \multicolumn{3}{c}{83.2 / 64.3 / 35.0}          \\
\quad+Dual-Reflect   & \multicolumn{3}{c}{83.5 / 70.4 / 44.2}                                                     & \multicolumn{3}{c}{77.9 / 57.1 / 31.3}                                                     & \multicolumn{3}{c}{74.0 / 62.0 / 14.2}          & 70.3 / 58.6 / 10.4          & 71.5 / 54.2 / 13.4                         & \multicolumn{3}{c}{83.2 / 65.3 / 35.1}          \\
\quad+IBUT   & \multicolumn{3}{c}{\textbf{84.0} / 70.7 / \textbf{44.6}} & \multicolumn{3}{c}{\textbf{79.2} / 58.9 / \textbf{31.8}} & \multicolumn{3}{c}{\textbf{75.0 / 64.3 / 19.3}} & \textbf{73.4 / 60.9 / 12.2} & \textbf{73.4 / 55.4 / 14.6}                & \multicolumn{3}{c}{\textbf{84.2 / 66.2 / 35.7}} \\ \hline
\end{tabular}}
\caption{The main results from the cultural MT dataset are presented. The bold indicates the highest values that are statistically significant, with p-values less than 0.01 in the paired t-test against all compared methods.}

\label{culturedata}
\end{table*}

\textbf{Comparative Methods.} In our evaluation, IBUT is compared with a range of translation methods, including Zero-shot \cite{DBLP:conf/iclr/WeiBZGYLDDL22}, 5-shot \cite{brown2020language}, Rerank \cite{DBLP:conf/eamt/MoslemHKW23}, Refine \cite{chen2023iterative}, MAD \cite{liang2023encouraging}, TEaR~\cite{DBLP:journals/corr/abs-2402-16379}, Dual-Reflect~\cite{chen-etal-2024-dual}, and MAPS \cite{He2023ExploringHT}. To validate its generalizability, we utilize three LLMs, which include closed-source models such as ChatGPT~\cite{ouyang2022training} and GPT-4~\cite{achiam2023gpt} \footnote{The ChatGPT and GPT-4 models used in this work are accessed through the gpt-3.5-turbo and gpt-4 APIs, respectively.}, as well as open-source models like Alpaca-7B~\cite{alpaca} \footnote{https://huggingface.co/tatsu-lab/alpaca-7b-wdiff/tree/main}, Vicuna-7B~\cite{vicuna2023} \footnote{https://huggingface.co/lmsys/vicuna-7b-v1.5}, and Qwen2.5-7B\cite{qwen2.5}~\footnote{https://modelscope.cn/models/Qwen/Qwen2.5-7B-Instruct/summary}. Details on comparative methods are in Appendix \ref{sec:comparative_methods}.

\textbf{Evaluation Metrics.} In evaluating our translation methodology, we initially employ COMET\footnote{https://huggingface.co/Unbabel/wmt22-comet-da} \cite{rei-etal-2022-comet-reference} and BLEURT\footnote{https://github.com/lucadiliello/bleurt-pytorch} \cite{sellam-etal-2020-bleurt-report} as automatic metrics, aligning with the established standards in LLM-based translation literature \cite{He2023ExploringHT,huang2024aligning}. For traditional translation evaluation, we use BLEU \footnote{https://github.com/mjpost/sacrebleu} \cite{papineni2002bleu}. To further evaluate our translation method, we employ human evaluations to verify translation performance. Details on human evaluations are in Appendix \ref{sec:human_evaluation}.

\section{Experimental Results}
\label{Experimental Results}
\subsection{Main Results}

\textbf{The effectiveness of IBUT in general news translation tasks}. In the WMT22 general news tasks, as shown in Table \ref{wmt22test} (WMT23 results in the Appendix \ref{wmt23_results}), IBUT outperforms other methods across 13 language pairs and 3 evaluation metrics. Specifically, in the news domain, the IBUT method outperforms translations directly based on contextual understanding by +1.5 COMET and +1.4 BLEURT. This indicates that the IBUT method alleviates the issue of Understanding Distortion in the news domain.

\textbf{The effectiveness of IBUT in low-resource tasks.} We selected all low-resource tasks (Uk$\leftrightarrow$Cs, Ru$\leftrightarrow$Sah, Liv$\leftrightarrow$En, En$\rightarrow$Hr) from WMT22. As observed in Table \ref{wmt22test}, current low-resource tasks still pose challenges to LLMs. However, compared to baseline methods, IBUT achieved an average improvement of +2.6 COMET in these low-resource tasks, with increases of +4 and +6.5 COMET for Liv$\leftrightarrow$En, respectively.

\textbf{IBUT is effective across different language similarities.} In WMT22, we validated the IBUT model using tasks with different language similarities. Specifically, Uk$\leftrightarrow$Cs represents closely related languages; En$\rightarrow$De and En$\rightarrow$Hr are from the same language family; Liv$\leftrightarrow$En, Ru$\leftrightarrow$Sah, and En$\rightarrow$Ja are categorized as distant language families. The experimental results, as shown in Table \ref{wmt22test}, demonstrate significant improvements across different language similarities due to IBUT. Notably, for the selected distant family languages, there was an average increase of +3.4 COMT, highlighting IBUT's potential to enhance translation tasks in distant language families.

\subsection{Cross-domain generalizability of IBUT}

\textbf{IBUT Adapts to Cultural MT.} As shown in Table \ref{culturedata}, IBUT outperforms other methods across all 6 language pairs. For translation corpora containing cultural-specific items, the IBUT method achieved an average increase of +2.02 and +1.6 COMET compared to the ChatGPT and MAPS methods. Notably, in the En$\rightarrow$Ta translation task, IBUT outperformed ChatGPT by +5.5 COMET. The experimental results above indicate that IBUT is suitable for translation tasks in the cultural domain.
% \vspace{-0.05cm}

\textbf{IBUT performed well in commonsense translation tasks}. As shown in Table \ref{commenmt}, IBUT significantly outperformed other methods in commonsense MT tasks, achieving the best translation performance. Compared to the MAPS method, IBUT improved by +2 in the COMET metric, demonstrating an enhanced ability to generate higher-quality contextual understanding. Moreover, IBUT surpassed the MAD method, which relies on multi-agent debate feedback, showing its outstanding feedback quality. Notably, in translation tasks involving logical reasoning, IBUT's performance was even better than GPT-4, fully showcasing its exceptional reasoning ability.

\begin{table}[!ht]\small \centering
\scalebox{0.9}{ 
\begin{tabular}{lc}
\hline
    \multicolumn{1}{l}{\textbf{Commonsense}} & \textbf{Zh$\rightarrow$En}           \\ \cline{2-2}
    \rowcolor{gray!30} 
    \multicolumn{1}{c}{}                         & COMET $\uparrow$ /BLEURT $\uparrow$ /BLEU $\uparrow$ \\ \hline
    \multicolumn{1}{l}{GPT4}                                         & 82.0 / 71.0 / 32.6    \\ \hline
    \multicolumn{1}{l}{ChatGPT}                                    & 79.7 / 68.2 / 29.8    \\
   \quad+5-shot                                     & 79.6 / 68.5 / 28.7   \\
   \quad+Rerank                                     & 80.9 / 69.1 / 29.9    \\
    \quad+MAPS                                       & 81.9 / 69.4 / 27.2    \\
    \quad+Refine                                     & 81.3 / 69.0 / 28.1    \\
    \quad+MAD                                        & 82.0 / 70.8 / 29.1    \\
    \quad+Dual-Reflect                                        & 82.2 / 71.8 / 28.4    \\
    \quad+TEaR                                        & 81.5 / 68.3 / 28.4    \\
    \quad+IBUT                                       &\textbf{ 83.9 / 72.7 / 32.6}                \\ \hline
\end{tabular}}
\caption{The main results from the Commonsense MT benchmark are presented. The bold indicates the highest values, statistically significant with p-values less than 0.01 in the paired t-test against compared methods.}
\label{commenmt}
\end{table} 
\vspace{-0.5cm}

% \section{Analysis}
% We conduct a detailed analysis of the effectiveness of our experiment, primarily focusing on results from CommonsenseMT Zh$\rightarrow$En, unless otherwise specified.
\subsection{Automated Evaluation of Understanding Distortion and Translation Performance}

\textbf{\begin{figure*}[!th]
% \vspace{-0.8 cm}
\centering
\includegraphics[scale=0.0658]{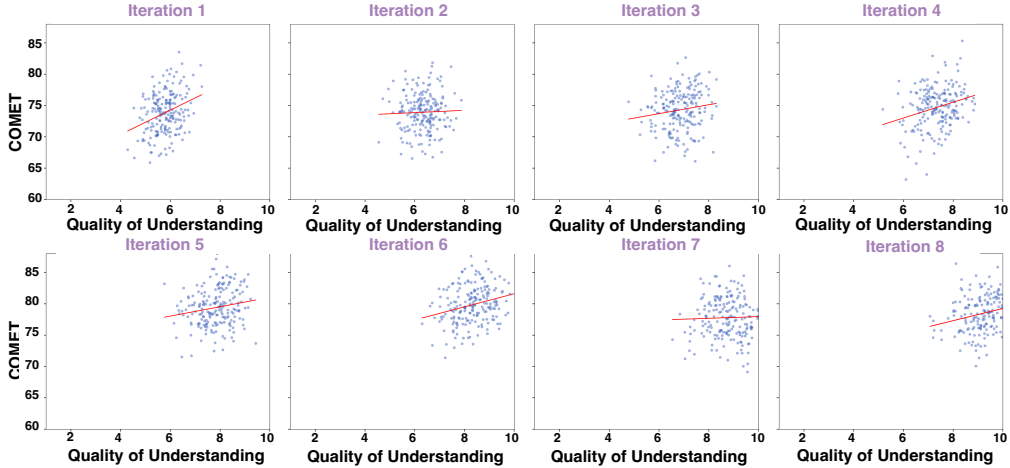} 
\caption{The experiment measures the relationship between the improvement in contextual understanding quality and translation performance during iterative refinement. }
\label{gpt4 scores}
\end{figure*}}

This study explored the positive impact of reducing understanding distortion issues in bilingual contextual understanding on translation performance using IBUT. We randomly selected a set of 200 Chinese$\rightarrow$English translation sentence pairs from the Commonsense MT dataset, which provides a test subset for lexical ambiguity. Based on the subset, IBUT iterated 8 times ($max\_iter=8$), saving the results of bilingual contextual understanding and translation COMET scores after each iteration. 

As shown in Figure \ref{gpt4 scores}, the vertical axis represents the translation performance, measured as the COMET score. The horizontal axis represents the scores evaluated by GPT-4 for the quality of bilingual contextual understanding affected by understanding distortion issues, with a maximum score of 10. The score for the source language is \(v_s\) and for the target language is \(v_t\), while the overall score $v$ is the average of the two (\(v=\frac{v_s + v_t}{2}\)). Details on the evaluation prompt can be found in Appendix \ref{gpt-4 evalution}.

The experimental results, as shown in Figure \ref{gpt4 scores}, demonstrate a positive correlation between the quality of contextual understanding and translation performance. Additionally, as the number of iterations increases, the quality of contextual understanding progressively improves, indicating that the IBUT method effectively reduces understanding distortion issues.

\subsection{Impact of Iterative Refinement on Translation Performance}

To further verify the impact of the Iterative Refinement part on overall translation performance, we conducted experiments on Cultural MT (En$\rightarrow$Zh) and Commonsense MT (Zh$\rightarrow$En), comparing methods like MAD and Refine to iteratively enhance translation quality. We set the maximum number of iterations at 9 and required that each iteration in the Iterative Refinement part obtain a new translation COMET score, rather than allowing adaptive termination in the Alignment Judgment part. 

\begin{figure}[!th]
\centering
\includegraphics[scale=0.0165]{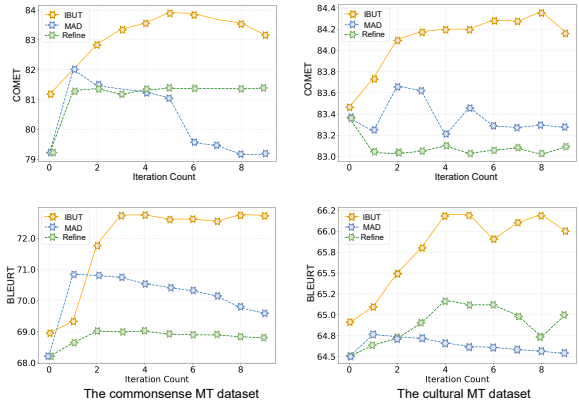} 
\caption{Analysis of the experimental setup for assessing the impact of the Iterative Refinement part on translation performance.}
\label{iter}
\end{figure}

The experimental results, as shown in Figure \ref{iter}, first indicate that IBUT surpasses the comparative methods in translation performance in most iterations, further proving the effectiveness of the method. Secondly, compared to the comparative methods, IBUT progressively enhances its performance in each iteration, demonstrating that the dual learning of translation can provide positive supervision signals in each iteration. 

\vspace{-0.3cm}
% \vspace{z}

To illustrate this iterative refinement more clearly, Table \ref{tab:translation-iterations} (in Appendix \ref{refinemet_case}) presents three cases where translations were correctly refined after a single iteration. These examples highlight how bilingual supervision signals contribute to enhancing translation quality through iteration.

\subsection{Human Evaluation}
\label{human_eval_error}
% 说明这个生成的内容对推理性的内容有强有力的性能（下面内容的表达需要构建一下）
\textbf{Human Evaluation of Understanding Distortion Issue.} In the human evaluation of understanding distortion issue, this study follows the method of \citealp{huang2024aligning} and \citealp{chen-etal-2024-dual} to assess translation outcomes from two main dimensions: accuracy in ambiguity resolution (commonsense domain) and the statistical results of understanding distortion issue (see Appendix \ref{sec:human_evaluation} for experimental setup details). 

The experimental results are shown in Table \ref{under_de}. Understanding distortion issues accounts for a significant proportion (40\%). Our method (IBUT) significantly addressed these failures, with a success rate of approximately 89\%, demonstrating the effectiveness of our method. Additionally, in terms of ambiguity resolution accuracy, IBUT outperformed the baseline by 13 acc points, indicating that bilingual understanding and iterative refinement contribute to enhancing ambiguity resolution capabilities in MT tasks.

\begin{table}[!ht]\small \centering
\begin{tabular}{@{}lcc@{}}
\toprule
\textbf{Methods}                   & \multicolumn{2}{c}{Human Evaluation} \\

\cmidrule(l){2-3} 
                                   &
                                   
                                   \textbf{Nums}     & \textbf{ACC $\uparrow$}     \\ \midrule
Understanding Distortion of Baseline & 28 (40\%)         & 65.9             \\
Understanding Distortion of IBUT     & \textbf{3 (-89\%) }        & \textbf{78.7}             \\ \bottomrule
\end{tabular}
\caption{The human-annotated results of the Commonsense MT benchmark. Baseline refers to the MAPS method modified into the form shown in Figure \ref{intro}(a). In the baseline method, there are \underline{70} sentences with translation errors.}
\label{under_de}
\end{table}

To better understand the limitations of the IBUT methods, Table \ref{tab:translation_errors_case} presents three sentences where IBUT still made translation errors in this experiment and analyzes them through human-annotated. These negative examples show that accurate translation depends on the source and target language achieving correct understanding through multiple iterations. If the LLMs misunderstand complex sentences during these iterations, translation errors will occur.

\begin{table}[h!]
    % \centering
    \scalebox{0.46}{
    \renewcommand{\arraystretch}{1.5}
    \begin{tabular}{>{\centering\arraybackslash}m{1cm} >{\centering\arraybackslash}m{6cm} m{8cm}}
        \toprule
        \textbf{No.} & \textbf{Human-annotated} & \textbf{Examples: Source/Error Result/Reference} \\
        \midrule
        1 & Nuanced translation errors arise from a lack of deep cultural understanding, leading to the loss of core meaning. & \textit{Source:} \begin{CJK*}{UTF8}{gbsn}如果不用心，就治不好学。\end{CJK*}  
        \newline \textit{Error:} If you don't put in the effort, you won't be able to cure poor learning.  
        \newline \textit{Right:} If you don't study by heart, you can't do scholarly research. \\
        \midrule
        2 & Although LLMs grasp that "\begin{CJK*}{UTF8}{gbsn}贩卖\end{CJK*} " implies "inculcate," textual noise hinders correction of mistranslations. & \textit{Source:} \begin{CJK*}{UTF8}{gbsn}贩卖资产阶级的精神鸦片。\end{CJK*}  
        \newline \textit{Error:} Peddling the bourgeoisie's spiritual opium.
        \newline \textit{Right:} Inculcate the spiritual opium of the bourgeoisie. \\
        \midrule
        3 & Iterative translation struggles to understand the meaning of "\begin{CJK*}{UTF8}{gbsn}起火\end{CJK*}" in Chinese, leading to mistakes. & \textit{Source:} \begin{CJK*}{UTF8}{gbsn}你家别起火了，到我家吃饭吧。\end{CJK*}  
        \newline \textit{Error:} The young gallants are new-born bucks in chase of bunny  
        \newline \textit{Right:} Young ones are like rabbits, new to the hunt, Born in a thatch of grass, on sandy ground \\
        \bottomrule
    \end{tabular}}
        \caption{Translation Errors with Examples.}
    \label{tab:translation_errors_case}
\end{table}
\vspace{-0.5cm}

\textbf{Transaltion Quality.} In human evaluation of translation quality, this study adopted the method \cite{liang2023encouraging} to validate translation quality on both the En$\rightarrow$Zh and Zh$\rightarrow$En test sets of the Cultural MT and the Commonsense MT dataset (Appendix \ref{sec:human_evaluation} for experimental setup details).

\begin{figure}[!th]
\includegraphics[scale=0.10]{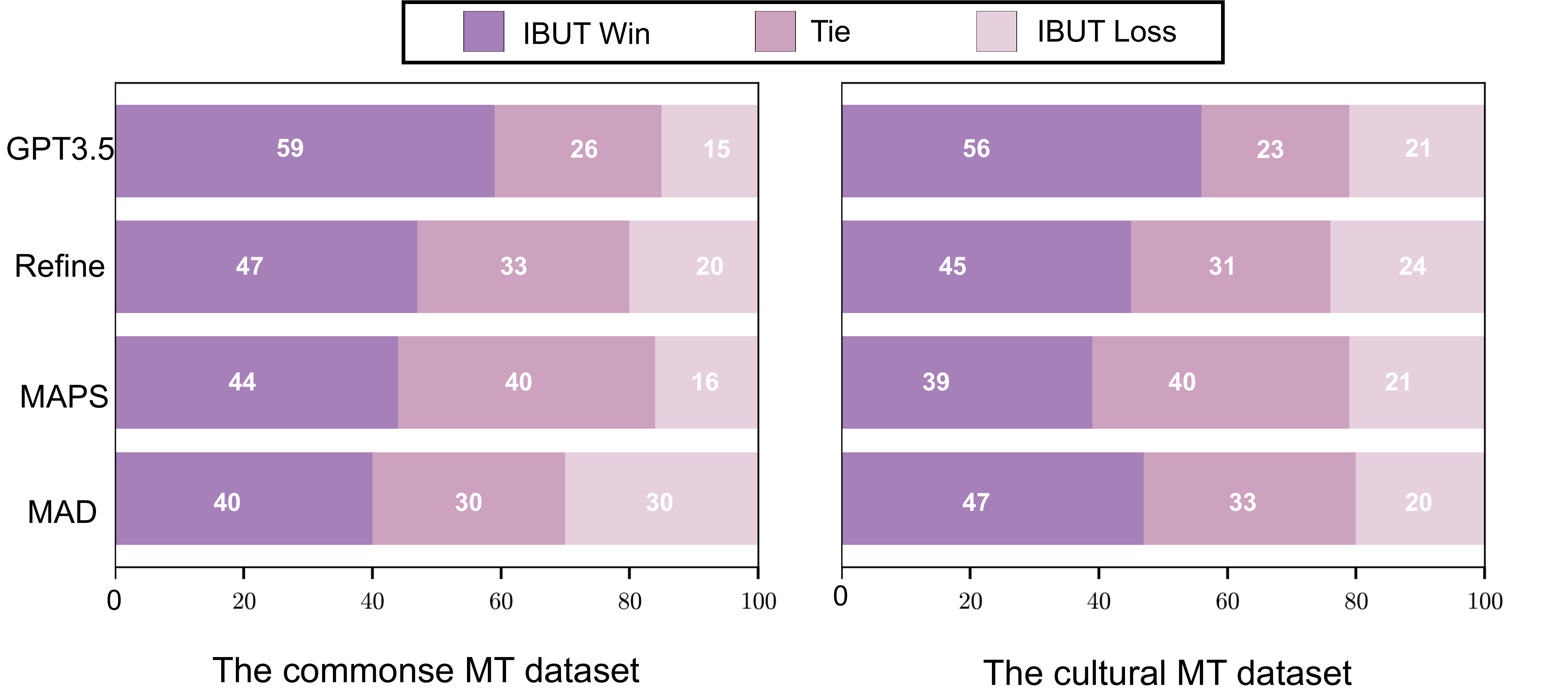} 
\caption{Human preference study comparing ChatGPT, Refine, MAPS, and MAD.}
\label{human_result}
\end{figure}

The experimental results are displayed in Figure \ref{human_result}. Within the Commonsense MT Dataset, IBUT performed best in terms of ambiguity resolution accuracy, thereby achieving higher human evaluation scores compared to other methods. In the Cultural MT Dataset, IBUT received higher human evaluation scores, indicating that its generated contextual understanding effectively enhances the performance of culturally translation tasks.

\vspace{-0.5cm}

% To further illustrate the process of IBUT’s iterative refinement of translation based on bilingual supervision signals, I selected three cases that were correctly translated after a single iteration for demonstration in tabel \ref{tab:translation-iterations} (in Appendix \ref{refinemet_case}).

\subsection{Effectiveness of Bilingual Contextual Understanding and Ablation Experiments}
\begin{figure}[!th]
\centering
\includegraphics[scale=0.027]{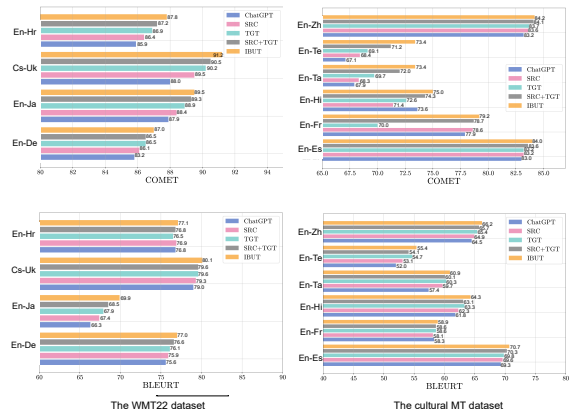} 
\caption{\textbf{Effectiveness of Bilingual Contextual Understanding and Ablation Experiments.} On the left are results for four language pairs from WMT22, and on the right are results for five language pairs from cultural MT. \textcolor{chatgpt_c}{ChatGPT} for direct translation; \textcolor{src_c}{SRC} for translation based on source language understanding; \textcolor{tgt_c}{TGT} for translation based on target language understanding; \textcolor{src_tgt_c}{SRC+TGT} for translation based on both source and target language understanding; \textcolor{ibut_c}{IBUT} as proposed method in section \ref{proposed_method}.}
\label{ablation}
\end{figure}
% \vspace{-0.5cm}

The IBUT introduced bilingual contextual understanding based on the source sentence to improve translation performance. To evaluate the effects of bilingual contextual understanding, we designed 5 control methods: (a) LLM-MT directly translating (\textcolor{chatgpt_c}{ChatGPT}); (b) LLM generating contextual understanding based on the source language, translated by LLM-MT (\textcolor{src_c}{SRC}); (c) LLM generating contextual understanding based on the target language, translated by LLM-MT (\textcolor{tgt_c}{TGT}); (d) LLM generating contextual understanding for both source and target languages, translated by LLM-MT (\textcolor{src_tgt_c}{SRC+TGT}); (e) using the IBUT method described in section \ref{proposed_method}.

\textbf{The effectiveness of Bilingual Contextual Understanding}. Figure \ref{ablation} shows that on WMT22 and cultural MT datasets, translation based on contextual understanding outperforms baseline methods, validating our research direction. Bilingual (SRC+TGT) contextual understanding notably improves performance over monolingual (SRC or TGT) understanding. Furthermore, target language (TGT) understanding has a greater impact on translation quality than source language (SRC) understanding.

\textbf{Ablation Experiments on IBUT Components}. Figure \ref{ablation} shows that using only the Understanding Generation component ("SRC or TGT") or skipping iterative refinement ("SRC+TGT") leads to inferior performance. These results validate the design rationale and effectiveness of the IBUT.

% As shown in Figure \ref{ablation}, "SRC or TGT" indicates the use of only the Understanding Generation part in IBUT, while "SRC+TGT" indicates that IBUT does not use iterative refinement. The experimental results demonstrate that the performance of the IBUT method surpasses these methods, thereby proving the rationality and effectiveness of the IBUT method design.

% . 

\subsection{IBUT Demonstrates Generalizability in Model Selection}
\label{general_model}

\begin{table}[!ht]\centering \small
\scalebox{0.85}{ 
\begin{tabular}{@{}lcllcll@{}}
\toprule
\textbf{WMT22}                                       & \multicolumn{3}{c}{\textbf{En$\rightarrow$De}}       & \multicolumn{3}{c}{\textbf{En$\rightarrow$Ja}}       \\ \midrule
\rowcolor{gray!30}  & \multicolumn{6}{c}{COMET $\uparrow$ /BLEURT $\uparrow$ /BLEU $\uparrow$}                                                     \\ \midrule
Alpaca-7B                                   & \multicolumn{3}{c}{75.5 / 62.2 / 11.3}          & \multicolumn{3}{c}{56.6 / 31.4 / 6.6}           \\
\quad+5shot                  & \multicolumn{3}{c}{76.3 / 62.8 / 12.1}          & \multicolumn{3}{c}{57.9 / 31.9 / 7.0}           \\
\quad+MAPS                   & \multicolumn{3}{c}{76.7 / 63.5 / 12.6}          & \multicolumn{3}{c}{58.3 / 33.9 / 7.5}           \\
\quad+IBUT                   & \multicolumn{3}{c}{\textbf{78.4 / 64.9 / 13.1}} & \multicolumn{3}{c}{\textbf{61.3 / 34.8 / 8.2}}  \\ \midrule
Vicuna-7B                                   & \multicolumn{3}{c}{79.8 / 67.4 / 15.2}          & \multicolumn{3}{c}{82.3 / 58.7 / 9.4}           \\
\quad+5shot                  & \multicolumn{3}{c}{80.3 / 67.8 / 15.3}          & \multicolumn{3}{c}{83.3 / 59.3 / 9.5}           \\
\quad+MAPS                   & \multicolumn{3}{c}{81.1 / 68.4 / 16.1}          & \multicolumn{3}{c}{84.4 / 60.3 / 9.8}           \\
\quad+IBUT                   & \multicolumn{3}{c}{\textbf{82.0 / 69.1 / 17.3}} & \multicolumn{3}{c}{\textbf{85.1 / 61.1 / 11.0}} \\ \midrule
Qwen2.5-7B                                   & \multicolumn{3}{c}{62.5 / 43.4 / 15.2}          & \multicolumn{3}{c}{64.5 / 31.7 / 7.1}           \\
\quad+5shot                  & \multicolumn{3}{c}{62.6 / 43.6 / 15.3}          & \multicolumn{3}{c}{64.0 / 31.7 / 7.3}           \\
\quad+MAPS                   & \multicolumn{3}{c}{62.3 / 43.3 / 15.2}          & \multicolumn{3}{c}{64.5 / 31.8 / 7.3}           \\
\quad+IBUT                   & \multicolumn{3}{c}{\textbf{63.2 / 44.7 / 16.0}} & \multicolumn{3}{c}{\textbf{66.1 / 33.0 / 9.2}} \\ 
\bottomrule
\end{tabular}}
\caption{The experimental results of IBUT on open-source models. The bold indicates the highest values that are statistically significant, with p-values less than 0.01 in the paired t-test against all compared methods.}
\label{wmt22test_general_open}
\end{table}
\vspace{-0.5cm}
To validate the generalizability of the IBUT method on open-source models, we selected two open-source models (Alpaca and Vicuna) for experimental verification. The experimental results, as shown in Table \ref{wmt22test_general_open}, indicate that the overall performance trends of the two open-source models are consistent with those observed using the GPT3.5 model. This demonstrates the generalizability of the IBUT method in open-source models. Additionally, we further validated the effectiveness of the IBUT method in GPT-4. The results are shown in Appendix \ref{general_all}.

\subsection{Computational Resource Analysis}

Since the IBUT method requires multiple iterative steps, it is necessary to discuss and analyze its resource consumption. For token consumption, we used the gpt-3.5-turbo tokenizer\footnote{https://github.com/openai/tiktoken} to tokenize and then calculated the token consumption of the comparative methods requiring iteration on the commonsense dataset.

\begin{table}[!ht]\centering \footnotesize
\scalebox{0.8}{ 
\begin{tabular}{@{}lccc@{}}
\toprule
   \textbf{Methods}     & \textbf{Avg I/O $\downarrow$} & \textbf{COMET $\uparrow$ /BLEURT $\uparrow$ /BLEU $\uparrow$} \\ \midrule
ChatGPT & 11.7 / 24.4       & 79.7 / 68.2 / 29.8    \\
\quad+5-shot & 59.4 / 35.6       & 79.9 / 68.6 / 30.2    \\
\quad+MAPS   & 167.7 / 172.2      & 81.9 / 69.4 / 27.2    \\
\quad+MAD    & 202.2 / 224.4      & 82.0 / 70.8 / 29.1    \\
\quad+IBUT   & 194.6 / 209.3      & 83.9 / 72.7 / 32.6    \\ \bottomrule
\end{tabular}}
\caption{
The statistics of methods inference cost on the commonsense dataset. The I/O represent Input/Output.
}
\label{token_nums}
\end{table}
\vspace{-0.5cm}
Table \ref{token_nums} shows that the IBUT method increases token consumption by 5 times compared to the 5-shot method, yet achieves substantial performance gains in COMET/BLEURT/BLEU metrics (+4.0/+4.1/+2.4). IBUT performs comparably to strong methods like MAD and MAPS, with an average improvement of 2 points. The computational trade-offs of long-context processing and inference time are detailed in Appendix \ref{long_com_reusult} and Appendix \ref{Computational Costs}, respectively. This limitation is discussed in the Limitations section as a future research direction for MT.

% Additionally, we discuss this limitation in the Limitations section, noting it as a future research direction for the MT community. 

\section{Conclusion}
% In this paper, we introduce a new method called \textbf{I}terative \textbf{B}ilingual \textbf{U}nderstanding \textbf{T}ranslation (IBUT), designed to address the inherent challenges of decreased translation performance in LLM-MT due to error introduced by contextual understanding. The IBUT initially generates bilingual contextual understanding, and then constructs a supervisory signal based on the dual learning of the translation task, thereby iteratively refining bilingual contextual understanding to enhance LLM-MT performance. The method performs well in various scenarios involving general news translation tasks, commonsense MT, and cultural MT datasets, with human evaluations confirming its effectiveness.
This paper presents \textbf{I}terative \textbf{B}ilingual \textbf{U}nderstanding \textbf{T}ranslation (IBUT), a method for improving LLM-based machine translation (LLM-MT) by addressing Understanding Distortion issue. IBUT generates bilingual contextual understanding, uses dual learning to create a supervisory signal, and iteratively refines the understanding to enhance translation performance. The method shows strong results across general news, commonsense, and cultural MT tasks, with human evaluations validating its effectiveness.

\section{Limitations}
\label{limitations}
The IBUT method has several limitations. Firstly, models with stronger understanding and generation capabilities will obtain better contextual understanding, thereby enhancing translation performance. Additionally, since our method requires multiple steps, it necessitates a significant amount of computational resources.

\bibliography{anthology,custom}

\clearpage

\appendix

\section{Experiment Setup}
\label{sec:Experiment_Setup}
\subsection{Detailed prompt for part-1}
\label{detail stage1}

\begin{mdframed}[backgroundcolor=purple!10, linecolor=white, linewidth=2pt, roundcorner=10pt]\small
\textbf{Part-1: Understanding Generation:} Please fully understand the meaning of the following $L^s$ text from your memory and describe your  understanding of key concepts, definitions, examples, and explanations of specific terms related to the translation task in $L^s$/$L^t$:$s$

\textbf{Input Text}: \begin{mdframed}[backgroundcolor=blue!10, linecolor=purple!10, linewidth=2pt, roundcorner=10pt]Source Sentence \(s\)\end{mdframed}\small

\textbf{Output Text}: \begin{mdframed}[backgroundcolor=yellow!10, linecolor=purple!10, linewidth=2pt, roundcorner=10pt]$C_s$ or $C_t$\end{mdframed}\small
\end{mdframed}

% \newpage

\subsection{Detailed prompt for part-2}
\label{detail stage2}

\begin{mdframed}[backgroundcolor=purple!10, linecolor=white, linewidth=2pt, roundcorner=10pt]\small
\textbf{Part-2: Alignment Judgment-1:} If you are a $L^s$ and $L^t$ linguist, determine whether provided source contextual understanding $C_s$ and target contextual understanding $C_t$, based on the source sentence $s$, convey different key concepts, definitions, examples, and explanations of specific terms related to the translation task. If so, provide a '$True$' response; otherwise, give a '$False$' response.

\textbf{Input Text}: \begin{mdframed}[backgroundcolor=blue!10, linecolor=purple!10, linewidth=2pt, roundcorner=10pt]Source Sentence \(s\) and source/target contextual understanding $C_s$/$C_t$\end{mdframed}\small

\textbf{Output Text}: \begin{mdframed}[backgroundcolor=yellow!10, linecolor=purple!10, linewidth=2pt, roundcorner=10pt]$True$ or $False$\end{mdframed}\small
\end{mdframed}

% \newpage

\begin{mdframed}[backgroundcolor=purple!10, linecolor=white, linewidth=2pt, roundcorner=10pt]\small
\textbf{Part-2: Alignment Judgment-2:} If you are a linguist proficient in both $L^s$ and $L^t$, based on the core meaning of the source sentence $s$, analyze the source contextual understanding $C_s$ / the target contextual understanding $C_t$. Generate verbal feedback in the language of $C_s$/$C_t$ to correct any current errors in $C_s$/$C_t$.

\textbf{Input Text}: \begin{mdframed}[backgroundcolor=blue!10, linecolor=purple!10, linewidth=2pt, roundcorner=10pt]Source Sentence \(s\), source/target language understanding $C_s$/$C_t$\end{mdframed}\small

\textbf{Output Text}: \begin{mdframed}[backgroundcolor=yellow!10, linecolor=purple!10, linewidth=2pt, roundcorner=10pt]$F_s$ or $F_t$\end{mdframed}\small
\end{mdframed}

        % background_rp = f'Please fully understand the meaning of the following {src_lang} text from your memory and describe your  understanding of key concepts, definitions, examples, and explanations of specific terms related to the translation task in {src_lang}:\n{src_text}'

\newpage

\subsection{Detailed prompt for part-3}
\label{detail stage3}

\begin{mdframed}[backgroundcolor=purple!10, linecolor=white, linewidth=2pt, roundcorner=10pt]\small
\textbf{Part-3: Iterative Refinement:} If you are a linguist proficient in both $L^s$ and $L^t$, based on the core meaning of the source sentence s and the opinions from $F_s$/$F_t$, further modify the current $C_t$/$C_s$.

\textbf{Input Text}: \begin{mdframed}[backgroundcolor=blue!10, linecolor=purple!10, linewidth=2pt, roundcorner=10pt]Source Sentence \(s\), source/target contextual understanding $C_s$/$C_t$ and source/target verbal feedback $F_s$/$F_t$\end{mdframed}\small

\textbf{Output Text}: \begin{mdframed}[backgroundcolor=yellow!10, linecolor=purple!10, linewidth=2pt, roundcorner=10pt]$C_s$ or $C_t$\end{mdframed}\small
\end{mdframed}

\subsection{Detailed prompt for part-4}
\label{detail stage4}

\begin{mdframed}[backgroundcolor=purple!10, linecolor=white, linewidth=2pt, roundcorner=10pt]\small
\textbf{Part-4:Understanding-Based Translation:} Based on $C_t$ and $C_s$, translate the following text from $L^s$ to $L^t$.

\textbf{Input Text}: \begin{mdframed}[backgroundcolor=blue!10, linecolor=purple!10, linewidth=2pt, roundcorner=10pt]Source Sentence \(s\), source/target contextual understanding $C_s$/$C_t$\end{mdframed}\small

\textbf{Output Text}: \begin{mdframed}[backgroundcolor=yellow!10, linecolor=purple!10, linewidth=2pt, roundcorner=10pt]Target Sentence $t$\end{mdframed}\small
\end{mdframed}

\subsection{Dataset Detail}
\label{data_appendix}

For the WMT22 test set \cite{kocmi-wmt22-etal-2022-findings}, the experimental analysis covers 9 language pairs. We used the full test dataset. Among these languages, Sah$\leftrightarrow$Ru, Uk$\leftrightarrow$Cs, En$\rightarrow$Hr and En$\leftrightarrow$Liv are classified as low-resource languages, respectively. 

For the WMT23 test set \cite{kocmi-etal-2023-findings}, the experimental analysis covers 4 language pairs. We used the full test dataset. Among them, En$\rightarrow$De and En$\rightarrow$Ja are identified as high and medium-resource languages, with the former belonging to the same language family and the latter exhibiting significant differences. 

The Commonsense MT dataset \cite{he-etal-2020-box-emnlp} encompasses vocabulary that requires common knowledge for resolution, along with instances of ambiguity in Zh$\rightarrow$En translation data. Each translation data includes a source sentence and two contrasting translations, involving seven different types of common knowledge. Although these sentences appear suitable for direct translation, they often lead to misleading interpretations.

The cultural MT dataset \cite{DBLP:journals/corr/abs-2305-14328} introduces a culturally relevant parallel corpus, enriched with annotations of cultural-specific items. This dataset encompasses 6 language pairs: En$\rightarrow$Es, En$\rightarrow$Fr, En$\rightarrow$Hr, En$\rightarrow$Ta, En$\rightarrow$Te, and En$\rightarrow$Zh. It also encompasses over 7,000 cultural-specific items from 18 concept categories across more than 140 countries and regions.

\subsection{Comparative Methods}
\label{sec:comparative_methods}

The following content will provide detailed descriptions of these comparative methods: 
%(§3.5) See Appendix A for specific details.

\begin{itemize}

\item \textbf{Baseline} is standard zero-shot translation performed in ChatGPT \cite{ouyang2022training} and GPT-4 \cite{achiam2023gpt}. The temperature parameter set to 0, which is the default value for our experiments.

\item \textbf{5-Shot} \cite{hendy2023good} involves prepending five high-quality labelled examples from the training data to the test input.

\item \textbf{Rerank} \cite{DBLP:conf/eamt/MoslemHKW23} was conducted with the identical prompt as the baseline, employing a temperature of 0.3 \citep{moslem2023adaptive}. Three random samples were generated and combined with the baseline to yield four candidates. The best candidate was chosen through GPT-4.

\item \textbf{Refine \cite{chen2023iterative}} first requests a translation from ChatGPT, then provides the source text and translation results, and obtains a refined translation through multiple rounds of modifications.

\item \textbf{MAPS \cite{He2023ExploringHT}} incorporate the knowledge of keywords, topic words, and demonstrations similar to the given source sentence to enhance the translation process.

\item \textbf{Dual-Reflect \cite{chen2023iterative}} provide supervisory signals for large models to reflect on translation results through dual learning, hereby iteratively improving translation performance (the maximum number of iterations is set to 5).

\item \textbf{TEaR \cite{chen2023iterative}} propose the first systematic and effective LLM-based self-refinement translation framework. 

\item \textbf{MAD \cite{liang2023encouraging}} enhance the capabilities of large language models (LLMs) by encouraging divergent thinking. In this method, multiple agents engage in a debate, while an agent oversees the process to derive a final solution.

\item \textbf{IBUT} is proposed method in Sec.\ref{proposed_method}. The method uses only ChatGPT with a max number of iterations set to 8 ($max\_iter=8$).

\end{itemize}

% \subsection{Comparative Methods}
% \label{sec:Comparative_Methods}
% The following sections provide detailed descriptions of these comparisons:
% %(§3.5)

% \textbf{Baseline}, standard zero-shot translation is performed in ChatGPT \cite{ouyang2022training} and GPT-4 \cite{achiam2023gpt} with the temperature parameter set to 0, which is the default value for our experiments.

% \textbf{Rerank} was conducted with the identical prompt as the baseline, employing a temperature of 0.3, in alignment with \citealp{moslem2023adaptive}. Three random samples were generated and combined with the baseline to yield four candidates. The optimal candidate was chosen through GPT-4. 

% \textbf{Renfie \cite{chen2023iterative}} first requests a translation from ChatGPT, then provides the source text and translation results, and obtains a refined translation through multiple rounds of modifications by mimicking the human correction process. 

% \textbf{MAPS \cite{He2023ExploringHT}}, incorporating the knowledge of keywords, topic words, and demonstrations similar to the given source sentence to enhance the translation process, respectively.

% \textbf{MAD \cite{liang2023encouraging}}, incorporating the knowledge of keywords, topic words, and demonstrations similar to the given source sentence to enhance the translation process, respectively.

\section{Experiment Results}

\subsection{Performance and Overhead of Long-Context Processing}
\label{long_com_reusult}
In the commonsense test datasets, the benchmark includes only one bilingual meaning word per sentence to better evaluate performance. To further analyze the performance and computational overhead of complex long-context processing, we concatenated $N$ sentences from the commonsense test datasets to create longer sentences. For instance, $N = 3$ means three source sentences are combined into one longer sentence. We then evaluated this modified dataset, and the results are shown in Table \ref{tab:performance_comparison}.

\begin{table}[ht]
\centering
\scalebox{0.8}{
\begin{tabular}{@{}lcc@{}}
\toprule
\textbf{Method}             & \textbf{Avg I/O $\downarrow$} & \textbf{COMET $\uparrow$ /BLEURT $\uparrow$ /BLEU $\uparrow$} \\ \midrule
\multicolumn{3}{c}{\textbf{$N = 2$}}                                          \\ \midrule
ChatGPT                     & 28.7 / 72.0      & 72.2 / 61.4 / 23.8         \\
\quad+MAPS & 351.5 / 407.1      & 76.9 / 66.3 / 26.1         \\
\quad+MAD    & 433.4 / 624.1      & 78.4 / 67.1 / 25.6         \\
\quad+IBUT   & 456.8 /  613.0     & 80.4 / 68.9 / 27.4         \\ \midrule
\multicolumn{3}{c}{\textbf{$N = 3$}}                                          \\ \midrule
ChatGPT                     & 37.9 / 57.7      & 72.1 / 60.2 / 22.7         \\
\quad+MAPS & 481.4 / 499.7      & 75.1 / 66.2 / 25.4         \\
\quad+MAD    & 610.9 / 675.4      & 77.2 / 66.2 / 25.8         \\
\quad+IBUT   & 510.3 / 609.2      & 78.6 / 67.8 / 27.2         \\ \bottomrule
\end{tabular}}
\caption{Evaluation Results for Different Methods with $N=2$ and $N=3$}
\label{tab:performance_comparison}
\end{table}
The experimental results demonstrate that IBUT outperforms both direct translation by LLMs and other multi-step LLM-MT methods, even when handling longer sentences containing multiple bilingual meaning words. For more complex and lengthy sentences, IBUT's computational overhead increases significantly due to the need to generate more concepts or terms. However, its translation performance remains superior. Therefore, developing more efficient and resource-efficient methods is an important direction for future research.

\subsection{Computational Costs}
\label{Computational Costs}
We illustrate with our method based on Vicuna-7B, using a single A100 GPU with 80G. Our proposed IBUT method has an inference speed of 6.71s/sample with a batch size of 2 and memory usage of 17657MiB. If using Vicuna-7B for zero-shot inference, under the same batch size settings, the inference speed is 4.72s/sample with memory usage of 14965MiB.

\subsection{The Experiment Setting of Error Reduction and Translation Enhancement }
\label{gpt-4 evalution}

For the Commonsense MT lexical ambiguity subset, first manually annotate the correct understanding of ambiguous words. The annotated data includes the source language Chinese and the target language English. Details of the scoring prompt for GPT-4, focusing on the reduction of error in bilingual contextual understanding after iterative refinement, are as follows:

\begin{mdframed}[backgroundcolor=purple!10, linecolor=white, linewidth=2pt, roundcorner=10pt]
\small
\textbf{Prompt for GPT-4 Evaluation} Please evaluate the source input $s$, contextual understanding $C_s$/$C_t$, and the manually annotated meanings of lexical ambiguities to assess if the contextual understanding includes error content to the translation. 

Scoring Guide:

1-2 points: The contextual understanding completely deviates from the source input, leading to generated content that is severely incorrect or irrelevant.

3-4 points: The contextual understanding partially deviates from the source input, resulting in partially relevant content with evident issues.

5-6 points: Although the contextual understanding does not completely deviate, there are errors in the interpretation of the source input, leading to content that is partially correct but flawed.

7-8 points: The contextual understanding is fundamentally accurate, correctly handles the source input, and the generated content is largely correct with only minor errors.

9-10 points: The contextual understanding is completely accurate, perfectly handles the source input and lexical ambiguities, and the generated content fully meets the requirements, successfully avoiding irrelevant content.

Based on these guidelines, score the model response from 0 to 10. Provide only the total score (just a number), without scores or explanations for each aspect. The score is \_\_.

\textbf{Input Text}: \begin{mdframed}[backgroundcolor=blue!10, linecolor=purple!10, linewidth=2pt, roundcorner=10pt]Source Sentence \(s\), source/target context understanding $C_s$/$C_t$\end{mdframed}

\textbf{Output Text}: \begin{mdframed}[backgroundcolor=yellow!10, linecolor=purple!10, linewidth=2pt, roundcorner=10pt] The score is \_\_ \end{mdframed}
\end{mdframed}

\subsection{Results on WMT23}

To further validate the generalizability of the method, we conducted experiments on the WMT23 test set. The experimental results are shown in Table \ref{wmt23_main}.

\label{wmt23_results}

\begin{table*}[!ht]\centering
\begin{tabular}{lclclclcl}
\hline
\textbf{WMT23} & \multicolumn{2}{c}{\textbf{En$\rightarrow$De}} & \multicolumn{2}{c}{\textbf{En$\rightarrow$Ja}} & \multicolumn{2}{c}{\textbf{En$\rightarrow$He}} & \multicolumn{2}{c}{\textbf{Cs$\rightarrow$Uk}} \\ \cline{2-9} 
\rowcolor{gray!30} 
\multicolumn{1}{c}{Metrics}                      
          & \multicolumn{8}{c}{COMET $\uparrow$ /BLEURT $\uparrow$ /BLEU $\uparrow$}                                                                                                                                                             \\ \hline

ChatGPT   & \multicolumn{2}{c}{83.5/69.1/39.7}             & \multicolumn{2}{c}{87.3/60.2/9.7}              & \multicolumn{2}{c}{82.1/69.3/22.3}             & \multicolumn{2}{c}{86.7/74.1/27.2}             \\
\quad+5shot    & \multicolumn{2}{c}{83.7/69.4/40.1}             & \multicolumn{2}{c}{87.8/61.5/10.1}             & \multicolumn{2}{c}{82.5/69.8/22.5}             & \multicolumn{2}{c}{87.3/74.5/27.5}             \\
\quad+MAD      & \multicolumn{2}{c}{83.9/70.3/41.6}             & \multicolumn{2}{c}{88.0/63.1/9.4}              & \multicolumn{2}{c}{82.9/70.0/24.0}             & \multicolumn{2}{c}{87.5/74.9/28.5}             \\
\quad+MAPS     & \multicolumn{2}{c}{83.6/69.9/42.1}             & \multicolumn{2}{c}{87.9/62.6/9.8}              & \multicolumn{2}{c}{82.5/69.3/23.1}             & \multicolumn{2}{c}{87.8/74.6/28.0}             \\
\quad+Refine   & \multicolumn{2}{c}{83.5/68.9/41.8}             & \multicolumn{2}{c}{87.6/62.4/10.8}             & \multicolumn{2}{c}{82.3/68.8/23.7}             & \multicolumn{2}{c}{87.3/74.1/28.3}             \\
\quad+IBUT     & \multicolumn{2}{c}{\textbf{84.3/71.8/42.6}}    & \multicolumn{2}{c}{\textbf{88.5/63.8}\textbf{/14.0}}             & \multicolumn{2}{c}{\textbf{83.1/72.1/24.9}}             & \multicolumn{2}{c}{\textbf{88.1/77.9/30.4}}             \\ \hline 
\end{tabular}
\caption{The main results from WMT23 are shown. The highest values are in bold, with p-values less than 0.01.}
\label{wmt23_main}
\end{table*}

\subsection{Results on Reference-free metric}
To further clarify the robustness of our evaluation, we incorporated COMET-KIWI\footnote{https://github.com/Unbabel/COMET} ~\cite{rei-etal-2022-cometkiwi-report}, a reference-free metric in the COMET series. The experimental results are shown in Table \ref{kiwi}.

\begin{table}[!ht] \centering
\scalebox{0.8}{
\begin{tabular}{lcccc}
\hline
\textbf{Methods} & \textbf{En-De} & \textbf{En-Ja} & \textbf{Cs-Uk} & \textbf{En-Hr} \\ \hline
ChatGPT          &                &                &                &                \\
\quad+Rerank          & 82.1           & 84.4           & 83.6           & 83.4             \\
\quad+MAPS    & 82.4           & 84.2           & 83.0           & 83.4          \\
\quad+MAD    & 82.0           & 83.7           & 83.6           & 83.3           \\
\quad+IBUT & \textbf{83.6}  & \textbf{84.7}  & \textbf{84.2}  & \textbf{83.8}  \\ \hline
\end{tabular}}
\caption{WMT22 evaluation results on COMET-KIWI metric.}

\label{kiwi}
\end{table}

These results demonstrate that our method still outperforms comparison methods in terms of COMET-KIWI scores, thereby further confirming the robustness of our evaluation.

\subsection{General Performance}
\label{general_all}
To demonstrate the generalizability of the method, we conducted experiments in Section \ref{general_model}, verifying that IBUT is effective not only on closed-source models but also on open-source models. Finally, since GPT-4 is an updated model of GPT-3.5, our method's effectiveness on GPT-3.5 theoretically implies effectiveness on GPT-4. To further illustrate this point, we conducted experiments on GPT-4 for commonsense MT. The experimental results are shown in Table \ref{gpt-4-gen}.

\begin{table}[]\centering
\begin{tabular}{@{}cc@{}}
\toprule
Methods & COMET $\uparrow$ /BLEURT $\uparrow$ /BLEU $\uparrow$ \\ \midrule
GPT-4   & 82.0/71.0/32.6    \\
\quad+5 shot & 82.3/71.5/32.9    \\
\quad+Rerank & 82.9/72.0/32.9    \\
\quad+IBUT   & 84.3/73.6/32.8    \\ \bottomrule
\end{tabular}
\caption{General Performance of general performance on commonsense MT}
\label{gpt-4-gen}
\end{table}

The experimental results demonstrate that our method achieves significant improvements when applied to GPT-4, thereby indicating the generalizability of our approach.

\subsection{Human Evaluations}
\label{sec:human_evaluation}
% In this section, we conduct human evaluation to measure translation quality. We assess coherence, fluency, and ambiguity resolution. Four english native speakers were invited to participate, and 50 samples were randomly selected from translations generated by different methods. For the content with Chinese ambiguity in Commonsense MT, I ensured the correctness of the source side understanding by confirming it with classmates whose native language is Chinese. For translation quality, each sentence was rated on a scale from 1 to 5, with 3 indicating a pass, 4 showing substantial consistency with the reference, and 5 being the highest score. The final score is the average of these four ratings. Additionally, in the CommonsenseMT task, the four experts scored each sample for ambiguity resolution against the reference, awarding 1 point for resolved and 0 points for unresolved.
\textbf{Human Evaluation of Understanding Distortion Issue.} In this section, we conduct a human evaluation to measure translation quality. We assess understanding distortion issues and ambiguity resolution. We invited one annotator to participate (a professional translator). 
The annotator first identifies and counts the sentences with ambiguity errors in the Baseline translation. Then, among these erroneous sentences, the annotator further filters and counts those where the errors are caused by contextual understanding. Finally, the annotator identifies and counts the sentences where the Baseline translation is incorrect but the IBUT translation is correct, and where the contextual understanding in the IBUT translation generates the correct sentence.
Additionally, in the CommonsenseMT task, the five experts scored each sample for ambiguity resolution against the reference, awarding 1 point for resolved and 0 points for unresolved.

\textbf{Human Evaluation of Translation Quality.} We conducted a human preference study on both the English-Chinese and Chinese-English test sets of the Cultural MT Datasets and the Commonsense MT Dataset. We invited one annotator to participate (a professional translator), and we randomly selected 100 translation results of the same source sentences generated by methods such as ChatGPT, Refine, MAPS, MAD, and IBUT. In terms of translation quality, the annotators compared the translation results of IBUT against other comparative methods. For the same source sentences, if IBUT's translation quality is superior, it is marked as \textcolor{win}{IBUT Win}; if the translation qualities are comparable, it is marked as \textcolor{tie}{Tie}; if the translation quality of other methods is better, it is marked as \textcolor{loss}{IBUT Loss}. We conducted three rounds of revisions on all evaluation results to increase the fairness of the assessments as much as possible. For the content with Chinese ambiguity in the commonsense  MT dataset, we ensured the correctness of the source side understanding by confirming it with classmates whose native language is Chinese.

\subsection{IBUT Demonstrates Generalizability on Low-Resource Languages}
\label{low_general}

% 为了进一步探索IBUT方法在开源模型是否能在翻译低资源任务有效。我们通过WMT23的低资源方向进行实验\footnote{https://www.statmt.org/wmt22/translation-task.html}. 实验结果如表N所示，实验结果可以表明我们的方法可以显著提升开源模型在低资翻译的性能，进而更好证明IBUT的通用性。
To further explore whether the IBUT method can be effective in low-resource translation tasks using open-source models, we conducted experiments on the low-resource directions of WMT23\footnote{https://www.statmt.org/wmt22/translation-task.html}. The experimental results are shown in Table \ref{open_model_low}, demonstrating that our method significantly improves the performance of open-source models in low-resource translation, thereby further validating the generalizability of IBUT.
% \usepackage{booktabs}
\begin{table}[!ht] \small \centering
\begin{tabular}{@{}lcllcll@{}}
\toprule
WMT22                                       & \multicolumn{3}{c}{Cs$\rightarrow$Uk}       & \multicolumn{3}{c}{En$\rightarrow$Hr}       \\ \midrule
\rowcolor{gray!30}  Metrics & \multicolumn{6}{c}{COMET $\uparrow$ /BLEURT $\uparrow$ /BLEU $\uparrow$ }                                                     \\ \midrule
Alpaca-7B                                   & \multicolumn{3}{c}{74.1/52.4/8.31}          & \multicolumn{3}{c}{65.9/53.2/8.1}           \\
\quad+5shot                  & \multicolumn{3}{c}{75.9/53.1/8.3}           & \multicolumn{3}{c}{67.9/53.6/8.3}           \\
\quad+MAPS                   & \multicolumn{3}{c}{76.3/53.7/9.2}           & \multicolumn{3}{c}{68.1/54.2/8.9}           \\
\quad+IBUT                   & \multicolumn{3}{c}{\textbf{77.9/54.3/9.5}}  & \multicolumn{3}{c}{\textbf{69.2/55.1/9.0}}  \\ \midrule
Vicuna-7B                                   & \multicolumn{3}{c}{74.9/57.8/10.5}          & \multicolumn{3}{c}{69.3/57.7/9.9}           \\
\quad+5shot                  & \multicolumn{3}{c}{76.3/58.3/10.9}          & \multicolumn{3}{c}{70.2/58.1/10.7}          \\
\quad+MAPS                   & \multicolumn{3}{c}{77.2/59.6/11.1}          & \multicolumn{3}{c}{71.1/58.8/11.6}          \\
\quad+IBUT                   & \multicolumn{3}{c}{\textbf{78.3/60.7/11.5}} & \multicolumn{3}{c}{\textbf{72.9/60.4/13.1}} \\ \bottomrule
\end{tabular}
\caption{The experimental low-resource results of IBUT on open-source models.  Alpaca-7B and Vicuna-7B mean to perform translation directly through Zero-Shot. The bold indicates the highest values that are statistically significant, with p-values less than 0.01 in the paired t-test against all compared methods.}
\label{open_model_low}
\end{table}
\vspace{-0.5cm}

\subsection{Introduce the Full Names of Languages.}
To better understand the experimental setup, we present the language codes and their corresponding full language names in Table \ref{laguage_codes}.
% \usepackage{booktabs}
% \usepackage{booktabs}
\begin{table}[!ht]
\begin{tabular}{@{}cc@{}}
\toprule
\multicolumn{1}{c|}{Language Codes}           & Full Name of Language Code \\ \midrule
\multicolumn{1}{c|}{En}  & English                    \\
\multicolumn{1}{c|}{JA}  & Japanese                   \\
\multicolumn{1}{c|}{Cs}  & Czech                      \\
\multicolumn{1}{c|}{Uk}  & Ukrainian                  \\
\multicolumn{1}{c|}{De}  & German                     \\
\multicolumn{1}{c|}{Hr}  & Croatian                   \\
\multicolumn{1}{c|}{Ru}  & Russian                    \\
\multicolumn{1}{c|}{Hi}  & Hindi                      \\
\multicolumn{1}{c|}{Ta}  & Tamil                      \\
\multicolumn{1}{c|}{Te}  & Telugu                     \\
\multicolumn{1}{c|}{Fr}  & French                     \\
\multicolumn{1}{c|}{Es}  & Spain                      \\
\multicolumn{1}{c|}{Sah} & Yakut                      \\
\multicolumn{1}{c|}{Liv} & Livonian                   \\ \bottomrule
\end{tabular}
\caption{The language names corresponding to the language codes.}
\label{laguage_codes}
\end{table}

\subsection{Iterative Translation Refinemet Case Studies}
\label{refinemet_case}

\begin{table*}[htbp]
\centering
\small
\scalebox{0.54}{ 
\renewcommand{\arraystretch}{1.3}
\setlength{\tabcolsep}{5pt}
\begin{tabular}{p{0.7cm} p{3.3cm} p{3.5cm} p{3cm} p{3.5cm} p{3.3cm} p{3.5cm} p{3cm} p{2cm}}
\toprule
\textbf{Case} & \textbf{Source Sentence} & \textbf{Iteration 0 Source Understanding} & \textbf{Translated Result} & \textbf{Iteration 0 Target Understanding} & \textbf{Verbal Source Feedback} & \textbf{New Source Understanding} & \textbf{New Translated Result} & \textbf{Reference} \\
\midrule
\textbf{1} & \begin{CJK*}{UTF8}{gbsn}一颗子弹报销一个敌人\end{CJK*} & \begin{CJK*}{UTF8}{gbsn}...报销理解为取出..实例为将某物从某处拿出来...\end{CJK*} \newline (\textcolor{gray}{Reimbursement is understood as taking something out, for instance, removing an object from a certain place.}) & One bullet takes out one enemy. & ...In military or colloquial contexts, \begin{CJK*}{UTF8}{gbsn}"报销"\end{CJK*} can be used as a euphemism for eliminating, neutralizing, or killing an enemy... & \begin{CJK*}{UTF8}{gbsn}..."报销"不能被笼统理解为"取出"。结合具体语境对其用法进行解释，"报销"在军事或口语中的引申义有时用于隐喻消灭敌人。...\end{CJK*} \newline (\textcolor{gray}{"Reimbursement" cannot be broadly understood as "taking out." Its usage should be explained in specific contexts. In military or colloquial language, the extended meaning of "reimbursement" is sometimes used metaphorically to imply eliminating the enemy.}) & \begin{CJK*}{UTF8}{gbsn}...报销"被用作隐喻，表示通过一颗子弹来消灭或击毙一个敌人。...\end{CJK*} \newline (\textcolor{gray}{"Reimbursement" is used metaphorically to signify eliminating or killing an enemy with a single bullet.}) & One bullet eliminates one enemy. & A bullet eliminates an enemy. \\
\midrule
\textbf{2} & \begin{CJK*}{UTF8}{gbsn}他擅长于援引未被发掘的人才。\end{CJK*} & \begin{CJK*}{UTF8}{gbsn}...援引表示利用某种资源或机会来获得好处或成功...\end{CJK*} \newline (\textcolor{gray}{Citation refers to utilizing a certain resource or opportunity to gain benefits or achieve success.}) & He excels at tapping into undiscovered talents. & ...\begin{CJK*}{UTF8}{gbsn}"援引"\end{CJK*} means suggesting someone for a position, role, or recognition... & \begin{CJK*}{UTF8}{gbsn}...援引"的本意是指引用、引证某些观点、事例。但是将"援引"解释为"利用某种资源或机会来获得好处或成功"符合上下文。...\end{CJK*} \newline (\textcolor{gray}{The original meaning of "citation" refers to quoting or referencing certain viewpoints or examples. However, interpreting "citation" as "utilizing certain resources or opportunities to gain benefits or achieve success" aligns with the context.}) & \begin{CJK*}{UTF8}{gbsn}...可以理解为提供支持或帮助...\end{CJK*} \newline (\textcolor{gray}{It can be understood as providing support or assistance.}) & He excels at recommending undiscovered talents for positions. & He is good at recommending undiscovered talent. \\
\midrule
\textbf{3} & \begin{CJK*}{UTF8}{gbsn}我们决不让祖国的江山变色。\end{CJK*} & \begin{CJK*}{UTF8}{gbsn}...变色在这个上下文中，更多的是指"改变现状"...\end{CJK*} \newline (\textcolor{gray}{In this context, "changing colors" refers more to "changing the status quo."}) & We will never allow our country's borders to change. & ...\begin{CJK*}{UTF8}{gbsn}"变色"\end{CJK*} symbolizes any form of alteration that could compromise the nation's stability and governance... & \begin{CJK*}{UTF8}{gbsn}...可以通过增加描述层次，突出"变色"引发的后果及其对国家稳定和治理的影响...\end{CJK*} \newline (\textcolor{gray}{By adding layers of description, the consequences triggered by "changing colors" and its impact on national stability and governance can be highlighted.}) & \begin{CJK*}{UTF8}{gbsn}...在此句中，结合上下文，将"变色"理解为对国家状态产生负面影响的转变...\end{CJK*} \newline (\textcolor{gray}{In this sentence, considering the context, "changing colors" is understood as a transformation that negatively impacts the state of the nation.}) & We will never allow our nation's condition to change for the worse. & We will never let the motherland's mountains and rivers change to the wrong direction. \\
\bottomrule
\end{tabular}}
\caption{Examples Demonstrating IBUT's Iterative refinement of Translation (Chinese to English) Based on Bilingual Supervision Signals. \textcolor{gray}{Gray text} indicates English annotations for the Chinese.}
\label{tab:translation-iterations}
\end{table*}

The table \ref{tab:translation-iterations} illustrates how errors in source-language contextual understanding are rectified through the integration of target-language contextual understanding. Experimental results demonstrate that target-language contextual understanding provides valuable target-language insights, serving as supervisory signals to address previous misunderstandings. This iterative process refines bilingual contextual understanding, leading to enhanced translation performance.

\end{document}